\documentclass[letterpaper, 10 pt, conference]{ieeeconf}  %

\IEEEoverridecommandlockouts                              %

\usepackage{graphics} %
\usepackage{epsfig} %
\usepackage{mathptmx} %
\usepackage{times} %
\usepackage{amsmath} %
\usepackage{amssymb}  %
\usepackage{booktabs}
\usepackage{multirow}
\usepackage{graphicx}
\usepackage{subcaption}
\usepackage{xcolor}
\usepackage{tabularx}
\usepackage{caption}
\usepackage{cite}
\usepackage{wrapfig}
\usepackage{dashrule}
\usepackage{cuted}
\usepackage{caption} 
\usepackage{booktabs}
\usepackage{array}
\usepackage[table]{xcolor}
\usepackage{hyperref}

\usepackage{booktabs}
\usepackage{array}      %
  
\newcolumntype{Y}{>{\centering\arraybackslash}X}

\newcolumntype{Z}{>{\columncolor{avgcol}\centering\arraybackslash}X}
\newcommand{\task}[1]{\texttt{#1}}

\newcommand{\hdr}[2]{#1 {\scriptsize\cite{#2}}}
\usepackage{booktabs}
\newcolumntype{L}{>{\raggedright\arraybackslash}X}
\newcommand{\hdri}[1]{#1\scriptsize}

\definecolor{avgcol}{RGB}{232,245,233}

\newlength{\panelht}
\newcolumntype{G}{
    >{\columncolor{avgcol}}c
}

\title{\LARGE \bf
Spatially Aware World Action Model via Geometric Latent Diffusion
}

\author{}
\author{Javier Alejandro Lopetegui Gonzalez, Paul Pacaud and Cordelia Schmid \\ \centering{\href{https://jlopetegui98.github.io/projects/sa_wam.html}{\color{blue}https://jlopetegui98.github.io/projects/sa\_wam.html}}
\thanks{The authors are with Inria and the Département d'Informatique
de l'École Normale Supérieure, PSL Research University in Paris,
75013 Paris, France (e-mail:
  {\tt\small javier-alejandro.lopetegui-gonzalez@inria.fr, paul.pacaud@inria.fr, cordelia.schmid@inria.fr})
}}

\begin{document}
\bstctlcite{IEEEexample:BSTcontrol}

\maketitle
\thispagestyle{empty}
\pagestyle{empty}

\begin{abstract}

World Action Models (WAMs) leverage the capabilities of large-scale
  pretrained video diffusion models to jointly predict future observations and
 actions, inheriting rich visual and physical priors from
  internet-scale video. This has made them a promising paradigm for robot policy
   learning, yet the prevailing models operate exclusively on RGB observations
  and do not leverage 3D information. To bridge this gap, we
  introduce a Spatially Aware World Action Model (SA-WAM), which repurposes a pretrained video model for joint action, RGB, and depth prediction, enabling 3D-aware world modeling and action prediction within a single diffusion backbone. We use a nonlinear encoding that maps the unbounded depth signal into the bounded input domain expected by the frozen VAE tokenizer. This allows us to reuse the tokenizer without 3D-specific fine-tuning, incorporating geometric information without sacrificing the pretrained priors. SA-WAM achieves state-of-the-art results on the RoboCasa and LIBERO-Plus benchmarks, while simultaneously improving future-state predictions. 
  Furthermore, SA-WAM outperforms strong baselines in real-world evaluation using a UR5 robotic arm, with strong gains in randomized environments. 
  We analyze the correlation between world model prediction quality
 and rollout success, providing insights into WAM performance and avenues for its improvement.
\end{abstract}

\section{Introduction}
\label{sec:introduction}

Vision-Language-Action (VLA) models have achieved impressive
generalization by leveraging large-scale pre-training on
vision, language and robot data
\cite{zitkovich2023rt,kim24openvla,ghosh2024octo,
black2025pi0,physicalintelligence2025pi05,
physicalintelligence2026pi07}. However, VLA models are typically initialized from Vision-Language Models trained predominantly on static image and text data. Video generative models, trained on dynamic sequences, appear as a potentially stronger prior for modeling temporal evolution, motion
and physical interactions relevant for robotic policies. Early approaches exploit these priors by generating future video observations and decoding actions through inverse dynamics
\cite{du2023unipi,liang2025videogenerators,liang2024dreamitate}.
Video-Action Models (VAMs) instead condition action prediction on
video latents from a pretrained backbone \cite{pai2025mimicvideo},
but typically require multi-stage video-and-action training. World Action Models (WAMs) or  Unified World Models (UWMs) unify video and action generation
in a shared generative architecture~\cite{li2025uva,zhu2025uwm,shen2025videovla,
kim2026cosmos,ye2026dreamzero}, showing that jointly learning actions and
future states can improve policy performance.

\begin{figure}[t]
    \centering
    \includegraphics[width=\linewidth]
    {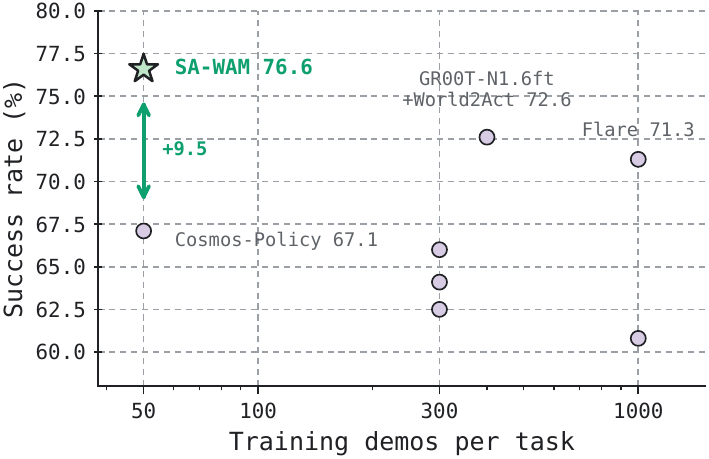}
\caption{\textbf{State-of-the-art results on the RoboCasa benchmark.} SA-WAM reaches $76.6\%$ success rate on RoboCasa with only 50 demonstrations
per task, improving over the matched Cosmos-Policy baseline by $9.5$
points and outperforming prior approaches trained with $6$--$20\times$
more data.}
\label{fig:data-efficiency}
\end{figure}

However, current WAMs mainly operate on RGB observations and do not leverage any 3D information. In contrast, 3D-aware VLA policies have
shown the value of 3D information in robotic policies, specifically for manipulation-intensive tasks where object geometry plays a critical role~\cite{zhen2024threedvla,qu2025spatialvla,li2025bridgevla, chen2026pointact,jia2026pointmappolicy}.
Bringing this spatial awareness to WAMs remains largely unexplored. 

\begin{figure*}[t]
    \centering
    \includegraphics[width=\textwidth]{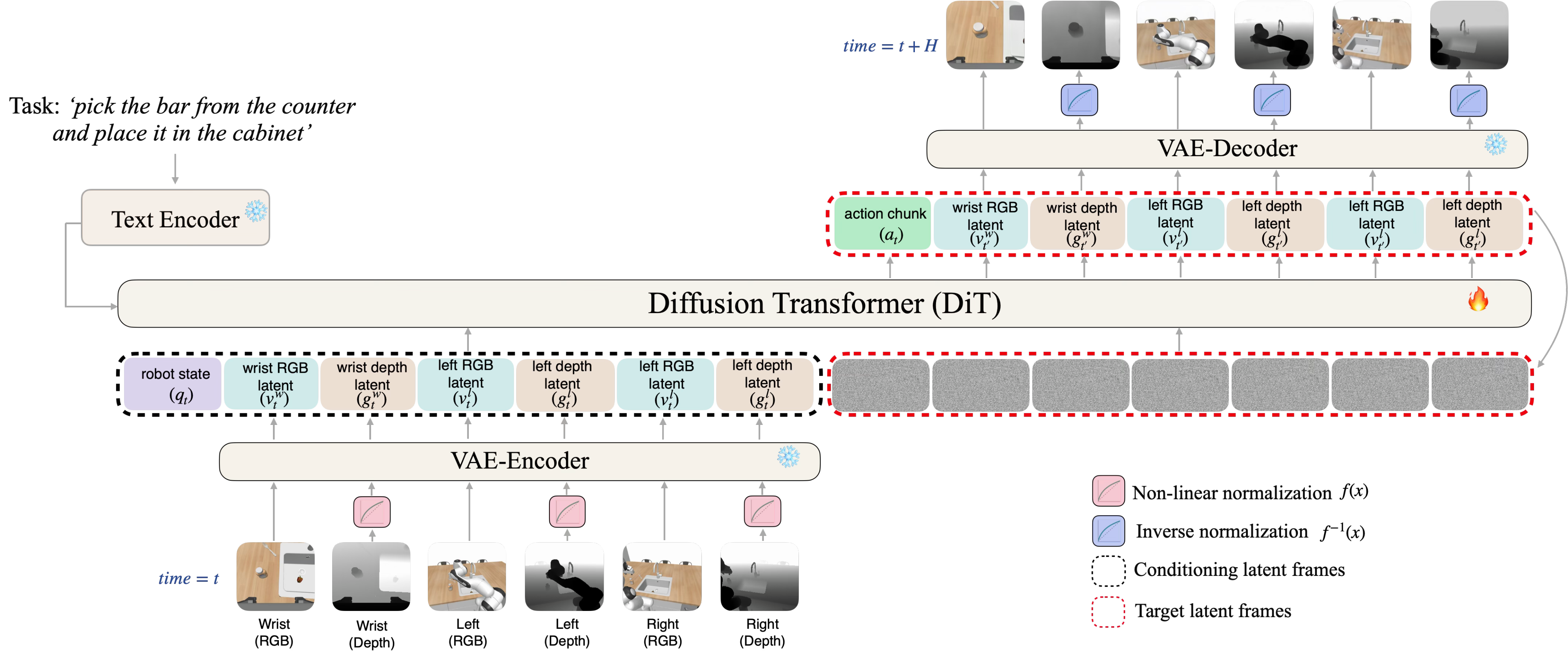}
    \caption{\textbf{SA-WAM architecture.}
    SA-WAM injects geometry into the WAM latent sequence as
    tokenizer-compatible frames paired with each RGB view, while the 
    proprioception ($q_t$) and action chunk ($a_t$) are inserted directly into dedicated positions in the latent sequence. The conditioning signals include the task description ($task$), one RGB frame per view ($v_t^c$) interleaved with its corresponding 3D modality ($g_t^c$) and the proprioceptive information at time $t$ ($q_t$). The action chunk, future-state RGB ($v_{t'}^c$), and future depth ($g_{t'}^c$) frames undergo the denoising process. The decoding phase is optional at inference time.}
    \label{fig:sawam_architecture}
\end{figure*}

We propose the Spatially Aware World Action Model (SA-WAM). By incorporating explicit geometry into world--action modeling, our approach significantly enhances policy performance; see Figure~\ref{fig:data-efficiency}. 
Following a latent-frame injection strategy
for adapting latent video diffusion models to visuomotor policy learning
\cite{kim2026cosmos}, we augment the Diffusion Transformer (DiT) latent sequence with depth signal (Figure~\ref{fig:sawam_architecture}). Depth is encoded as additional latent frames and processed by the same frozen pretrained VAE tokenizer as RGB observations. The key challenge is to encode this unbounded metric depth into the frozen VAE feature space while preserving geometric precision without using modality-specific 3D encoders. Motivated by nonlinear parameterizations used in
diffusion-based depth estimation
\cite{saxena2024dmd,gui2025depthfm}, SA-WAM employs nonlinear depth normalization.
The main intuition is to compress the far-field while preserving near-field  resolution, where
manipulation precision matters most.

\noindent Our main contributions are summarized as follows:\\
(i) 
We introduce SA-WAM, which integrates explicit 3D geometric information into World Action Models. We use nonlinear normalization of the depth to avoid training dedicated 3D encoders.\\
(ii) SA-WAM achieves state-of-the-art
results on RoboCasa~\cite{nasiriany2024robocasa} with only 50 demonstrations per task (Figure~\ref{fig:data-efficiency}) and on LIBERO-Plus~\cite{fei2025libero}. Moreover, it obtains strong real-world results on a UR5 robotic arm platform, demonstrating its sample efficiency and robustness to randomized environments.\\
(iii) We investigate different normalization strategies for depth, showing how log-scale normalization leads to stronger policy performance while remaining better matched to the frozen video tokenizer's distribution. Additionally, we analyze how geometric information improves WAM world modeling quality and study
the correlation between future prediction quality and policy rollout success.

\section{Related Work}
\label{sec:related_work}

\paragraph{World Models for Robot Policy Learning}
World models have long been used in robot learning to predict future states for planning and control
\cite{ha2018worldmodels,hafner2019planet}. More recent video-based robot policies leverage generative video models as learned priors for robot behavior. Methods such as UniPi~\cite{du2023unipi} and Dreamitate~\cite{liang2024dreamitate} synthesize future visual trajectories and recover the corresponding actions through inverse dynamics or action decoding. In contrast, Video-Action Models such as Mimic-Video~\cite{pai2025mimicvideo} augment a pretrained video backbone with a dedicated action-prediction head and follow a two-stage training procedure: first training the video model, then learning the action module conditioned on the video latent representations. More recent World Action Models (WAMs) and Unified World Models (UWMs), including VideoPolicy~\cite{liang2025videogenerators}, UWM~\cite{zhu2025uwm}, DreamZero~\cite{ye2026dreamzero}, Motus~\cite{bi2026motus}, and Cosmos-Policy~\cite{kim2026cosmos}, jointly model future states and actions within a shared generative architecture. Cosmos-Policy is particularly relevant to our work because it repurposes the latent sequence of a pretrained video diffusion transformer to represent robot-specific modalities, such as proprioception and actions. This design preserves the foundational video model architecture without introducing modality-specific modules, while exploiting video pretraining without explicitly generating the complete future video sequence. These methods demonstrate the value of visual foresight for policy learning, but they primarily operate on RGB observations.

\paragraph{3D-Aware Robot Policies}
Explicit 3D observations have been widely used to improve manipulation
policies by providing direct access to object geometry and spatial
relations. PolarNet~\cite{chen2023polarnet} predicts actions from
language-conditioned point-cloud representations, while
3D-VLA~\cite{zhen2024threedvla} connects 3D perception, reasoning, and
action through a generative world model. SpatialVLA~\cite{qu2025spatialvla}
introduces egocentric 3D positional encodings and adaptive spatial action
grids, whereas PointVLA~\cite{li2025pointvla} injects point-cloud
features into a pretrained VLA through a lightweight module.
PointACT~\cite{chen2026pointact} further couples hierarchical
point-cloud features with action tokens through multi-scale interaction.
PointMapPolicy~\cite{jia2026pointmappolicy} instead represents dense
3D coordinates as image-aligned PointMaps, allowing explicit geometry
to be processed using standard visual encoders. Similarly, SA-WAM uses a dense representation of the metric depth as its 3D modality and maps it into latent frames using the same frozen video tokenizer as RGB.

\paragraph{Metric Geometry Encoding for Pretrained Visual Models}
Geometric information can be represented through dedicated modality-specific latent spaces or adapted to the representation domain of existing visual models. DiffusionDepth~\cite{duan2024diffusiondepth} learns a compact depth-specific latent space using a lightweight autoencoder, enabling diffusion directly over depth representations. Alternatively, metric-depth methods commonly employ non-linear parameterizations to better distribute representational capacity across large depth ranges.
DMD~\cite{saxena2024dmd} uses log-scale depth to jointly model indoor
and outdoor scenes, while SwinMTL~\cite{taghavi2024swinmtl} applies
logarithmic scaling to emphasize the densely represented near-depth
range. DepthFM~\cite{gui2025depthfm} encodes log-normalized depth
within the bounded input range of a pretrained latent autoencoder,
showing improved performance over linear normalization.
GRIN~\cite{guizilini2025grin} similarly performs pixel-level diffusion
in log-depth space to improve precision across large distance ranges.
SA-WAM transfers this principle to WAMs: we encode depth using a nonlinear normalization that makes metric geometry compatible with a frozen
video tokenizer while allocating greater resolution to
manipulation-relevant regions.

\section{Method}
\label{sec:methodology}

\subsection{3D Modality Injection for World Action Modeling}
\label{sec:method_geometry_injection}

SA-WAM adapts a pretrained latent video diffusion model into a spatially
aware World Action Model that jointly predicts robot actions and
visual-geometric world evolution. Conditioned on the task instruction,
current proprioception, and multi-view RGB-D observations, the model
jointly denoises an action chunk together with the corresponding future
RGB-D observations. Following~\cite{kim2026cosmos}, robot
proprioception and actions occupy dedicated positions in the latent
sequence. In contrast, RGB and depth observations are encoded into
latent frames using the same frozen video tokenizer. This design
repurposes the pretrained video model for unified world and action
generation without introducing dedicated action heads or
geometry-specific encoders, as illustrated in
Figure~\ref{fig:sawam_architecture}.

Let $\mathcal{V}=\{w,l,r\}$ denote the wrist, left, and right cameras,
respectively. For each camera $c\in\mathcal{V}$, let $I_t^c$ denote the
RGB observation and $d_t^c$ its corresponding metric depth map at time
$t$. We denote the frozen VAE encoder by $E$, and define the RGB and
geometric latent frames as
\begin{equation}
    v_t^c = E(I_t^c),
    \qquad
    g_t^c = E(I_{d,t}^c),
    \label{eq:modality_latents}
\end{equation}
where $I_{d,t}^c$ is a tokenizer-compatible three-channel representation
of the metric depth $d_t^c$, defined below.

Let $\ell$ denote the task instruction, $q_t$ the robot proprioception,
and $a_t$ the action chunk starting at time $t$. The conditioning is
$\mathcal{C}_t=\left(\ell,q_t,\{v_t^c,g_t^c\}_{c\in\mathcal{V}}\right)$.
The model jointly denoises $a_t$ and the future observation latents
$\{v_{t'}^c,g_{t'}^c\}_{c\in\mathcal{V}}$ at $t'=t+H$, where $H$
denotes the prediction horizon. Thus, RGB and depth occupy separate
latent-frame positions while sharing the same pretrained tokenizer.

Reusing the frozen video tokenizer for depth requires mapping metric
depth, which is positive and unbounded in principle, to the bounded
input range expected by the VAE encoder. We therefore define a
camera-specific normalization $f$ that transforms metric depth into a
tokenizer-compatible image representation. We compare three choices:
\textit{linear}, \textit{inverse-depth}, and \textit{log-scale}
normalization.

\paragraph{Depth Injection}
For each camera $c$, let $d_t^c(u,v)>0$ denote the metric depth at pixel
$(u,v)$. Since the wrist and external cameras cover different depth
ranges, we estimate robust camera-specific bounds from the training
data. Let $\mathcal{D}_c$ contain all valid depth values observed by
camera $c$ across the training set, and define
$d_{\min}^c=P_2(\mathcal{D}_c)$ and
$d_{\max}^c=P_{98}(\mathcal{D}_c)$, where $P_p$ denotes the $p$-th
percentile. Each depth observation is then clamped as
$d_t^{c,\star}(u,v)=
\operatorname{clip}(d_t^c(u,v),d_{\min}^c,d_{\max}^c)$.

All depth encodings follow the same procedure. A monotone normalizer
$f(\cdot;a,b):[a,b]\rightarrow[0,1]$ is applied to the clamped metric
depth, after which its output is rescaled to the $[-1,1]$ tokenizer
input range and replicated across three channels:
\begin{equation}
\begin{aligned}
    \hat d_t^c(u,v)
    &= 2f\!\left(
        d_t^{c,\star}(u,v);
        d_{\min}^c,d_{\max}^c
    \right)-1, \\
    I_{d,t}^c(u,v)
    &= \bigl(
        \hat d_t^c(u,v),
        \hat d_t^c(u,v),
        \hat d_t^c(u,v)
    \bigr)\in[-1,1]^3.
\end{aligned}
\label{eq:depth_encoding}
\end{equation}
The resulting depth representation is encoded by the frozen VAE
according to ~\eqref{eq:modality_latents}.

The three encoding strategies differ only in the choice of $f$:
\begin{equation}
f(z;a,b)=
\begin{cases}
\dfrac{z-a}{b-a},
    & \text{linear},\\[6pt]
\dfrac{1/z-1/b}{1/a-1/b},
    & \text{inverse-depth},\\[6pt]
\dfrac{\log(z/a)}{\log(b/a)},
    & \text{log-scale}.
\end{cases}
\label{eq:depth_normalizations}
\end{equation}
Their local sensitivity $\left|\partial f/\partial z\right|$ is constant
for linear normalization, proportional to $1/z^2$ for inverse-depth
normalization, and proportional to $1/z$ for log-scale normalization.
Linear normalization therefore distributes representational resolution
uniformly across the metric range. Inverse-depth normalization
concentrates it strongly on nearby geometry, whereas log-scale
normalization provides an intermediate behavior, allocating greater
resolution to the near field while retaining useful precision at larger
distances. Each mapping is invertible over the retained depth interval,
allowing predictions to be mapped back to metric depth for evaluation.

\subsection{Training Objective}
\label{sec:method_training_objective}

We initialize the DiT backbone from the open-source Cosmos-Predict2 model~\cite{nvidia2025cosmospredict2}, a pretrained video model, specifically trained for physical AI, that provides strong dynamic priors for robot policy learning. We use the corresponding Wan2.1 spatiotemporal VAE tokenizer, which remains frozen during training. Let $x_0$ denote the clean target sequence that
contains the action chunk together with the tokenizer latents of the future RGB
and geometric frames. Conditioning information $\mathcal{C}_t$
includes the task instruction, current proprioception, and current
multi-view RGB-D observations. Following the EDM formulation~\cite{karras2022elucidating}, $x_0$ is
corrupted with Gaussian noise $n\sim\mathcal{N}(0,\sigma^2I)$, and the
denoiser is optimized with
$\mathcal{L}_{\mathrm{EDM}}
=\mathbb{E}\!\left[
\lambda(\sigma)
\lVert D_\theta(x_0+n;\sigma,\mathcal{C}_t)-x_0\rVert_2^2
\right]$,
where $\lambda(\sigma)$ is the noise-dependent weighting. The same
objective jointly supervises the action, future RGB and future depth
slots, allowing all modalities to be learned within a single generative
sequence without dedicated prediction heads or modality-specific losses. 

 \section{Simulation Experiments}
  \label{sec:experiments}
\subsection{Experimental setup} 
\label{sec:exp_benchmarks}

\paragraph{Simulation Benchmarks} For our evaluation we use three simulation benchmarks, RoboCasa \cite{nasiriany2024robocasa}, LIBERO \cite{liu2023libero} and LIBERO-Plus \cite{fei2025libero}.

RoboCasa~\cite{nasiriany2024robocasa} is a large-scale simulation benchmark for everyday kitchen
manipulation, built on RoboSuite/MuJoCo with a Franka Panda arm. We
evaluate on its 24 atomic tasks, including pick-and-place,
articulated-object interaction, appliance control, and faucet
manipulation. Each task provides two third-person cameras, a
wrist-mounted camera, and proprioception. We report average task
success rate over three random seeds.

LIBERO~\cite{liu2023libero} is a simulation benchmark for robotic manipulation using a Franka Panda  arm. We evaluate on its four standard suites: \textsc{Spatial},
\textsc{Object}, \textsc{Goal}, and \textsc{Long}. 
The first three
isolate spatial, object, and goal-conditioned generalization, while
\textsc{Long} evaluates longer-horizon task execution.  Furthermore, we evaluate on LIBERO-Plus~\cite{fei2025libero}, which
extends the same suites with controlled perturbations along seven
dimensions (object layout, camera viewpoint, robot initial state,
language instruction, lighting, background texture, and sensor noise). For fair comparison against baselines, we apply the same noise perturbation to depth as to RGB  when evaluating sensor noise. 
Following its official zero-shot protocol, we train only on the original LIBERO
demonstrations and evaluate directly on LIBERO-Plus, without any
adaptation to the perturbed conditions.

\paragraph{Training Details} We fine-tune the 2B-parameter Cosmos-Predict2 DiT and keep the
Wan2.1 VAE tokenizer frozen. For RoboCasa and LIBERO, training is run on 40 NVIDIA H100 GPUs
with an effective batch size of 800 and 960, respectively. We optimize the EDM objective with AdamW using a peak learning rate of
$10^{-4}$. After an initial warm-up, the learning rate is linearly
decayed until 30,000 steps, at which point it is reduced by an additional
factor of $5$ and kept constant for the remainder of training. Training
runs for 45,000 iterations on RoboCasa and 40,000 iterations on LIBERO. During training we sample noise using a hybrid approach, from the base CosmosPredict2~\cite{agarwal2025cosmos}
log-normal distribution $p_{\mathrm{base}}(\sigma)$ with probability $0.7$
and from $\mathcal{U}(1,85)$ with probability $0.3$~\cite{kim2026cosmos}. The model predicts action
chunks of length $H{=}32$, of which the first 16 steps are executed
open-loop.

\subsection{Ablations on the depth normalization approach}
\label{sec:exp_ablations}

We conduct all ablations on the same eight RoboCasa tasks:
\task{PnP\allowbreak Sink\allowbreak To\allowbreak Counter},
\task{PnP\allowbreak Counter\allowbreak To\allowbreak Sink},
\task{PnP\allowbreak Microwave\allowbreak To\allowbreak Counter},
\task{PnP\allowbreak Counter\allowbreak To\allowbreak Microwave},
\task{Open\allowbreak Drawer},
\task{Turn\allowbreak On\allowbreak Sink\allowbreak Faucet},
\task{Turn\allowbreak Off\allowbreak Sink\allowbreak Faucet} and
\task{Coffee\allowbreak Serve\allowbreak Mug}.
This subset spans pick-and-place, articulated-object interaction, faucet
manipulation, and mug-serving behaviors. We first analyze how different depth
encodings preserve metric information through the tokenizer and then evaluate
downstream policy performance.

\paragraph{Metric reconstruction analysis}
We first focus on understanding the impact of each normalization approach on the ability of the frozen tokenizer to handle the geometric representation. We evaluate the normalization approaches discussed in Section~\ref{sec:method_geometry_injection}: \textit{linear, inverse} and \textit{log-scale}. Given a depth observation, we perform a closed encode-decode round trip and evaluate the reconstructed depth and report the Absolute Relative Error (AbsRel) with respect to the input. The corresponding overall results for the three cameras and the depth-range breakdown for the wrist camera are reported in Table~\ref{tab:geometry_ablation_reconstruction}.

Linear and inverse normalization exhibit complementary failure modes. Linear
depth preserves medium- and far-range geometry well, but substantially loses
precision in the near field, reaching $6.68\%$ AbsRel below $0.3\ \mathrm{m}$.
Inverse-depth normalization shows the opposite behavior: it is highly accurate in the near field
($0.62\%$), but it degrades sharply at medium and far distances ($4.16\%$ and
$13.19\%$, respectively). Log normalization provides a more balanced profile,
achieving the lowest overall error ($0.63\%$) while
maintaining comparatively low error across the wrist-camera depth ranges.

\begin{table}[t]
  \centering
  \footnotesize
  \setlength{\tabcolsep}{5pt}

  \caption{Ablations on the eight-task RoboCasa subset.
    Metric depth reconstruction is reported as AbsRel
    (\%, $\downarrow$). Overall values aggregate all camera views,
    while range-specific values are reported for the wrist camera only.
    Policy success rate (SR, \%) is averaged over
    3 evaluation seeds $\times$ 50 trials per task.
    $\dagger$ denotes the configuration retained for subsequent
    experiments.}
  \label{tab:geometry_ablation}

  \begin{subtable}{\columnwidth}
    \centering
    \caption{Metric reconstruction}
    \label{tab:geometry_ablation_reconstruction}

    \begin{tabular}{@{}lcccc@{}}
      \toprule
      Normalization & Overall & ${<}0.3$\,m & $0.3$--$1$\,m & ${>}1$\,m \\
      \midrule
      Linear  & 1.47 & 6.68 & 0.80 & 1.78 \\
      Inverse & 1.41 & 0.62 & 4.16 & 13.19 \\
      Log     & 0.63 & 1.25 & 1.00 & 3.19 \\
      \bottomrule
    \end{tabular}
  \end{subtable}

  \vspace{10pt}

  \begin{subtable}{\columnwidth}
  \centering
  \caption{Policy-level ablation}
  \label{tab:geometry_ablation_policy}
  \begin{tabular}{@{}ccc}
    \toprule
    & Normalization  & SR (\%) \\
    \midrule
    RGB & -  & 52.7 \\
     RGB-D & Linear  & 56.5 \\
     RGB-D & Inverse  & 59.5 \\
     RGB-D & Log & \textbf{61.7}$^\dagger$ \\
    \bottomrule
  \end{tabular}
\end{subtable}
\end{table}

\paragraph{Policy-level ablation}
We train all configurations under the same protocol, using 50 demonstrations
per task and 6,000 gradient steps for faster iteration. As shown in Table~\ref{tab:geometry_ablation_policy}, all geometric approaches substantially improve over RGB-only method.
Log-scale depth normalization improves over linear normalization by $5.2$ points and over inverse normalization by $2.2$, consistent with the more
balanced metric precision observed above.

\begin{table*}[t]
\centering
\caption{
Category-wise success rate (SR, \%) on the 24 RoboCasa tasks.
The categories contain 8 pick-and-place tasks, 6 open/close tasks,
7 turn/toggle tasks, and 3 coffee tasks.
Category averages are computed from available per-task results. The final column reports the average
over all 24 tasks. Best overall results in \textbf{bold}.
}
\label{tab:robocasa_categories}

\small
\setlength{\tabcolsep}{5pt}
\renewcommand{\arraystretch}{1.08}

\begin{tabular*}{\textwidth}{
    @{\extracolsep{\fill}}
    lcccccG
}
\toprule
Method
& Train. demos / task
& PnP
& Open/Close
& Turn/Toggle
& Coffee
& Avg. \\
\midrule

$\pi_0$~\cite{black2025pi0}
& 300
& --
& --
& --
& --
& 62.5 \\

VideoPolicy~\cite{liang2025videogenerators}
& 300
& --
& --
& --
& --
& 66.0 \\

UWM~\cite{zhu2025uwm}
& 1000
& 35.6
& 82.0
& --
& --
& 60.8 \\

FLARE~\cite{zheng2025flare}
& 1000
& --
& --
& --
& --
& 71.3 \\

GR00T-N1.5~\cite{bjorck2025gr00t}
& 300
& 53.8
& 85.7
& 61.7
& 54.0
& 64.1 \\

GR00T-N1.5 + HAMLET~\cite{koo2026hamlet}
& 300
& 48.5
& 87.7
& 69.7
& 64.0
& 66.4 \\

GR00T-N1.6-ft + World2Act~\cite{dinh2026world2act}
& 300
& --
& --
& --
& --
& 72.6 \\

Cosmos-Policy~\cite{kim2026cosmos}
& 50
& 51.8
& 91.1
& 66.3
& 61.6
& 67.1 \\

\textbf{SA-WAM (ours)}
& 50
& 68.5
& 92.6
& 77.5
& 64.7
& \textbf{76.6} \\

\bottomrule
\end{tabular*}
\end{table*}

\begin{table*}[tbh]
\centering
\caption{Weighted-average success rate (SR, \%) on LIBERO-Plus under zero-shot
evaluation. Results are aggregated across the seven official perturbation
axes and weighted by the number of episodes in each axis. Best result in
\textbf{bold}.}
\label{tab:libero_main}
\footnotesize
\setlength{\tabcolsep}{3pt}
\setlength{\extrarowheight}{2pt}
\renewcommand{\tabularxcolumn}[1]{b{#1}}
\begin{tabularx}{\textwidth}{l YYYYYYY YY Z}
\toprule
Method
& \hdr{$\pi_0$}{black2025pi0}
& \hdr{$\pi_0$-FAST}{pertsch2025fast}
& \hdr{OpenVLA-OFT}{kim2025fine}
& \hdr{RIPT-VLA}{tan2025interactive}
& \hdr{$\pi_{0.5}$}{physicalintelligence2025pi05}
& \hdr{HoloBrain-0-GD}{lin2026holobrain}
& \hdr{ABot-M0}{yang2026abot}
& \hdr{Fast-WAM}{yuan2026fast}
& \hdr{Cosmos-Policy}{kim2026cosmos}
& \hdri{\textbf{SA-WAM (ours)}}{} \\
\midrule
Avg. SR
& 53.6 & 61.6 & 69.6 & 68.4 & 84.6 & 74.0 & 80.5
& 50.0 & 81.4 & \textbf{86.6} \\
\bottomrule
\end{tabularx}
\end{table*}

We therefore retain log-scale normalization for depth. It
achieves the highest policy success while providing balanced metric precision
across the workspace.
Unless stated otherwise, all subsequent experiments use this configuration,
denoted as SA-WAM.

\subsection{Comparison to the State of the Art}

In this section we compare SA-WAM to existing robot policy baselines on the RoboCasa and LIBERO-Plus benchmarks. 

\paragraph{RoboCasa results}

Table~\ref{tab:robocasa_categories} compares SA-WAM with strong VLA
policies~\cite{black2025pi0,bjorck2025gr00t}, video- and world-model
approaches~\cite{liang2025videogenerators,zhu2025uwm, kim2026cosmos}, and recent
future-modeling or post-training methods~\cite{zheng2025flare,
koo2026hamlet,dinh2026world2act}. We report the average success rate
across three seeds on the 24 RoboCasa tasks, together with the
category-level breakdown when available. SA-WAM achieves the highest
overall success rate at $76.6\%$ using only 50 demonstrations per task,
outperforming methods trained with substantially larger downstream data
budgets. In particular, compared to Cosmos-Policy~\cite{kim2026cosmos} under the same demonstration budget, SA-WAM improves by
$9.5$ points overall and performs consistently better across all task
categories. The largest improvements occur on PnP and Turn/Toggle tasks,
with gains of $16.7$ and $11.2$ points, respectively, indicating that
explicit geometric conditioning is particularly beneficial for
manipulations requiring accurate object localization and
spatial interaction. 

Figure~\ref{fig:robocasa_qualitative} shows one representative pick-and-place task where SA-WAM completes the manipulation,
whereas Cosmos-Policy fails. We observe that SA-WAM produces future predictions that
remain better aligned with the corresponding simulator rollout; see~\ref{sec:predict_versus_success} for a quantitative evaluation.

\paragraph{LIBERO and LIBERO-Plus results}
On standard LIBERO, SA-WAM achieves an average success rate of $98.4\%$,
remaining on par with the strongest policies on this largely saturated
benchmark. Table~\ref{tab:libero_main} compares SA-WAM with recent
VLAs~\cite{black2025pi0,pertsch2025fast,kim2025fine,
tan2025interactive,physicalintelligence2025pi05,lin2026holobrain,
yang2026abot} and World Action Models~\cite{yuan2026fast,
kim2026cosmos} on the more challenging LIBERO-Plus benchmark under
zero-shot evaluation. SA-WAM achieves the highest weighted-average
success rate at $86.6\%$, exceeding the strongest VLA baseline,
$\pi_{0.5}$, by $2.0$ points. Moreover, it improves over the
strong Cosmos-Policy baseline by $5.2$ points, indicating that explicit
geometric conditioning improves WAM robustness to perturbations. 

\subsection{World Modeling Quality Analysis}
\label{sec:exp_world_quality}

\begin{table*}[t]
\centering
\caption{World-model RGB prediction quality on RoboCasa.
F and W denote the averaged fixed-camera views and the wrist camera, respectively; subscript $m$ denotes masked region.
$\uparrow$ higher is better; $\downarrow$ lower is better.}
\label{tab:wm_quality}
\small
\setlength{\tabcolsep}{2pt}
\renewcommand{\arraystretch}{0.95}
\begin{tabularx}{\textwidth}{@{}l *{12}{>{\centering\arraybackslash}X}@{}}
\toprule
& \multicolumn{4}{c}{PSNR $\uparrow$}
& \multicolumn{4}{c}{SSIM $\uparrow$}
& \multicolumn{4}{c}{LPIPS $\downarrow$} \\
\cmidrule(lr){2-5} \cmidrule(lr){6-9} \cmidrule(lr){10-13}
Model
& F & W & F$_m$ & W$_m$
& F & W & F$_m$ & W$_m$
& F & W & F$_m$ & W$_m$ \\
\midrule
Cosmos-Policy
& 15.78 & 14.64 & 12.56 & 15.68
& .525 & .563 & .400 & .581
& .167 & .417 & .219 & .381 \\
SA-WAM
& \textbf{15.97} & \textbf{15.16} & \textbf{12.82} & \textbf{16.08}
& \textbf{.528} & \textbf{.566} & \textbf{.406} & \textbf{.585}
& \textbf{.161} & \textbf{.354} & \textbf{.207} & \textbf{.321} \\
\bottomrule
\end{tabularx}
\end{table*}

\begin{figure}[!t]
\centering
    \includegraphics[width=0.88\linewidth]{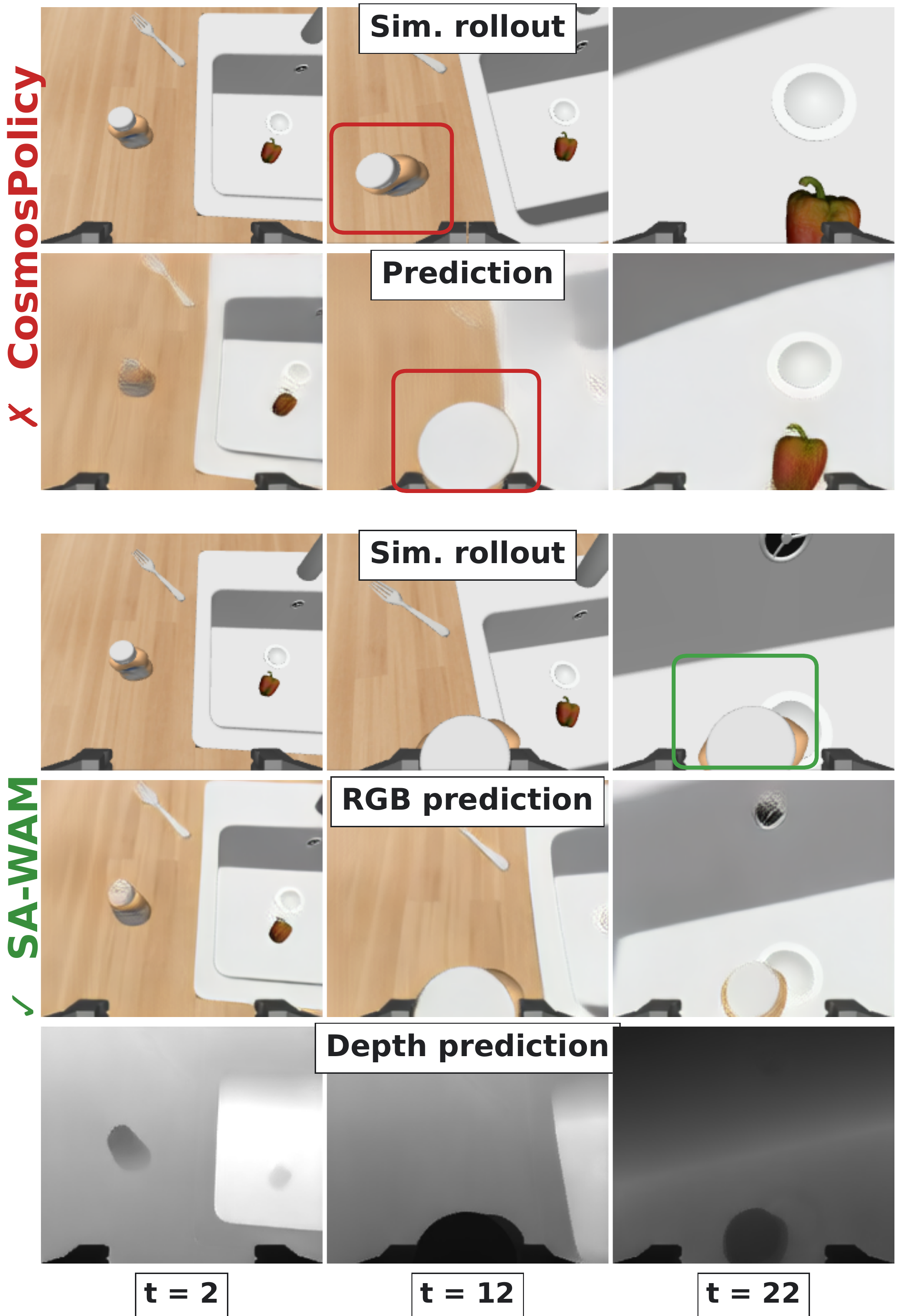}

\caption{\textbf{RoboCasa pick-and-place task:} \textit{Pick the condiment bottle from the counter and place it in the sink}. We show the wrist view before and during grasping, and at the final rollout stage. The upper block shows Cosmos-Policy and the lower block shows SA-WAM, with predictions compared against their corresponding simulator rollouts; for SA-WAM we also show depth predictions. Red boxes mark rollout inconsistencies, and the green box marks task completion. 
}
\label{fig:robocasa_qualitative}
\end{figure}
We analyze whether adding explicit geometry improves the visual
world-modeling quality of SA-WAM. Predicted future RGB frames are
compared with ground-truth simulator renders using
PSNR~\cite{huynh2008scope}, which measures pixel-level reconstruction
fidelity; SSIM~\cite{wang2004image}, which measures structural
similarity; and LPIPS~\cite{zhang2018unreasonable}, which measures
perceptual distance in a learned feature space. Higher PSNR and SSIM
indicate better predictions, whereas lower LPIPS is better. Metrics are averaged across the 24 RoboCasa tasks and reported separately for the fixed-camera views (averaged over the left and right cameras) and the wrist camera. We additionally report masked metrics computed over the
union of the robot arm, gripper, and target-object segmentation masks,
thereby isolating prediction quality in manipulation-relevant
regions~\cite{bardhan2026persistent}. As shown in Table~\ref{tab:wm_quality}, SA-WAM improves over
Cosmos-Policy across all reported metrics. The gains are most pronounced
for the dynamic wrist view, where close-range interaction and camera
motion make future prediction particularly challenging. Improvements also
persist under masked evaluation, indicating that they extend to the robot
and manipulated object rather than arising only from background or global
image statistics. 
The qualitative examples in Figure~\ref{fig:robocasa_qualitative}
complement these metrics: SA-WAM
produces RGB predictions that remain more consistent with its realized
rollout. Together with the corresponding gains in task success,
these results support the
hypothesis that, in WAM, stronger future-state modeling is associated with better
policy performance.

\subsection{Prediction Quality versus Rollout Success}
\label{sec:predict_versus_success}

\begin{figure}[t]
    \centering
    \begin{subfigure}[b]{\columnwidth}
      \includegraphics[width=0.9\linewidth]{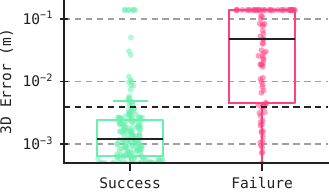}
      \caption{}\label{fig:corr_box}
    \end{subfigure}\\[-1pt]
  \vspace{1.em}
    
    \begin{subfigure}[b]{\columnwidth}
      \includegraphics[width=0.9\linewidth]{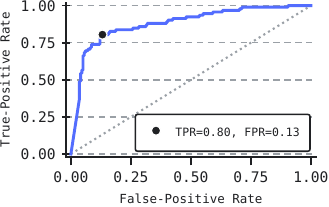}
      \caption{}\label{fig:corr_roc}
    \end{subfigure}
  \caption{\textbf{Future-prediction quality correlates with rollout success.}
  \textbf{(\ref{fig:corr_box})} Object-centric 3D prediction error is lower
  for successful rollouts.
  \textbf{(\ref{fig:corr_roc})} A threshold on this error detects failures
  with AUC~$\approx0.88$ and an operating point of $80\%$ detection at
  $13\%$ false positive rate.} 
  \label{fig:correlation}
\end{figure}

\begin{figure*}[!htb]
\centering
\begin{subfigure}{0.45\linewidth}
  \centering
  \includegraphics[width=\linewidth]
  {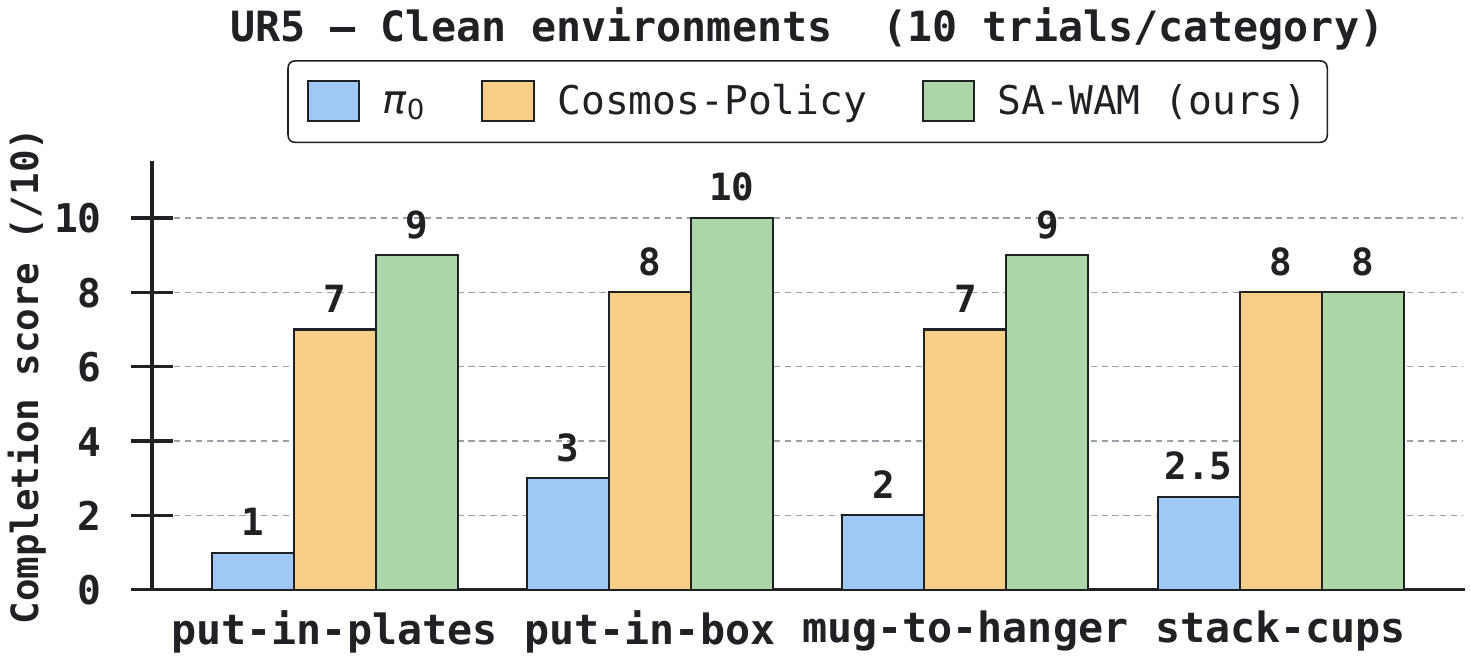}
  \caption{Clean environments}
  \label{fig:ur5_clean}
\end{subfigure}
\hfill
\begin{subfigure}{0.45\linewidth}
  \centering
  \includegraphics[width=\linewidth]
  {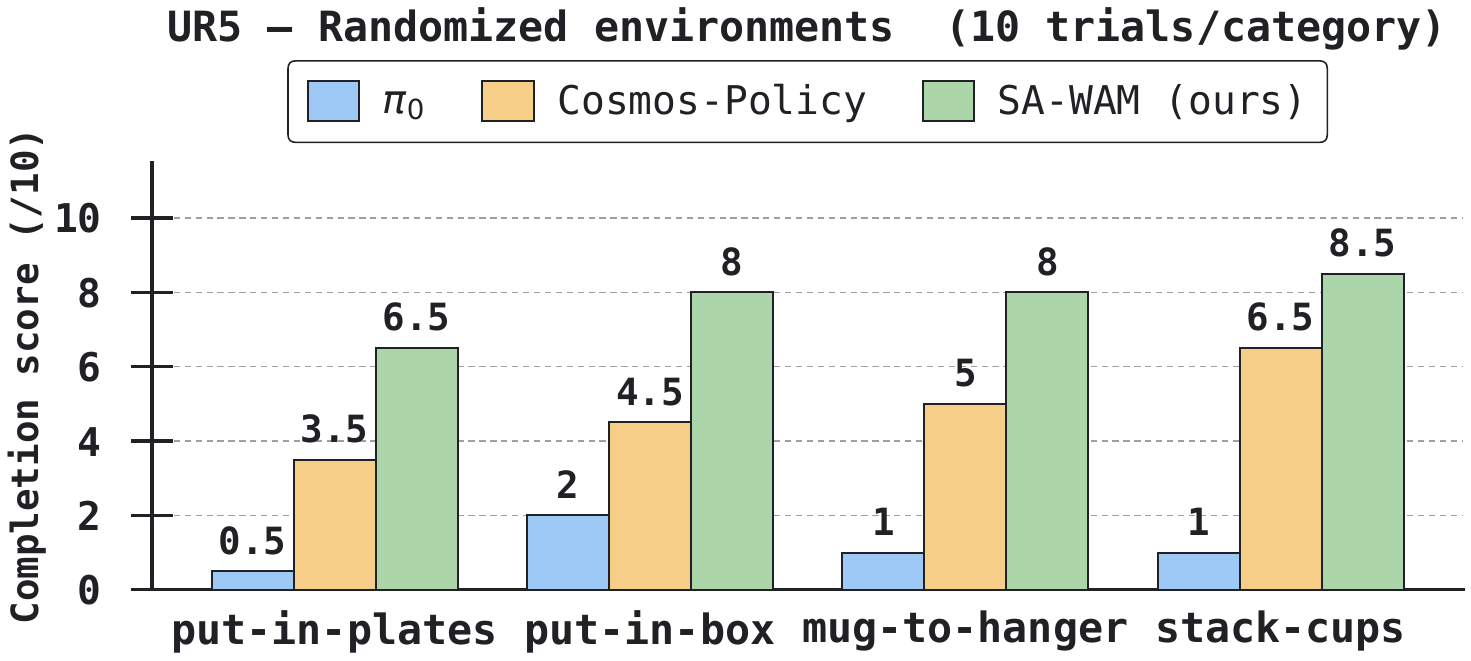}
  \caption{Randomized environments}
  \label{fig:ur5_rand}
\end{subfigure}
\caption{\textbf{Real-world UR5 results.}
Task-completion scores over 10 trials per category for $\pi_0$,
Cosmos-Policy, and SA-WAM in the clean and randomized settings.
SA-WAM retains the strongest performance under randomization, consistent
with depth providing a complementary spatial cue in the presence of
visually confusing distractors.}
\label{fig:ur5_results}
\end{figure*}

Next, we quantitatively study the correlation between future prediction consistency and rollout outcome. We focus on the manipulation-heavy pick-and-place (PnP) tasks for this analysis. We leverage the future depth prediction in SA-WAM to calculate an object-centric
geometric error from the wrist camera by comparing the
model-predicted future with the realized simulator rollout at a fixed
short horizon after the grasp attempt. Specifically, we use the
ground-truth simulator mask of the target object, back-project the
predicted and ground-truth depth values within this mask into 3D, and
measure the resulting object-region 3D prediction error. Figures~\ref{fig:correlation}(\subref{fig:corr_box}) and (\subref{fig:corr_roc}) show that this object-level error is strongly associated with rollout outcome. Successful rollouts tend to keep the predicted object aligned with the realized object, whereas failed rollouts show larger divergence around the manipulation event; see Figure~\ref{fig:correlation}(\subref{fig:corr_box}). Across PnP rollouts, successful episodes have lower 3D prediction
error than failed ones; a threshold on this error yields AUC~$\approx0.88$, detecting $80\%$ of failures at a $13\%$
false-alarm rate as we can see in Figure~\ref{fig:correlation}(\subref{fig:corr_roc}). These results suggest that task-relevant geometric future-prediction error can serve as a quantitative diagnostic of WAM failure modes.

\section{Real-World Experiments}
\label{sec:rw_experiments}
\begin{figure}[!tb]
\centering
\begin{subfigure}{\columnwidth}
\includegraphics[width=\linewidth]{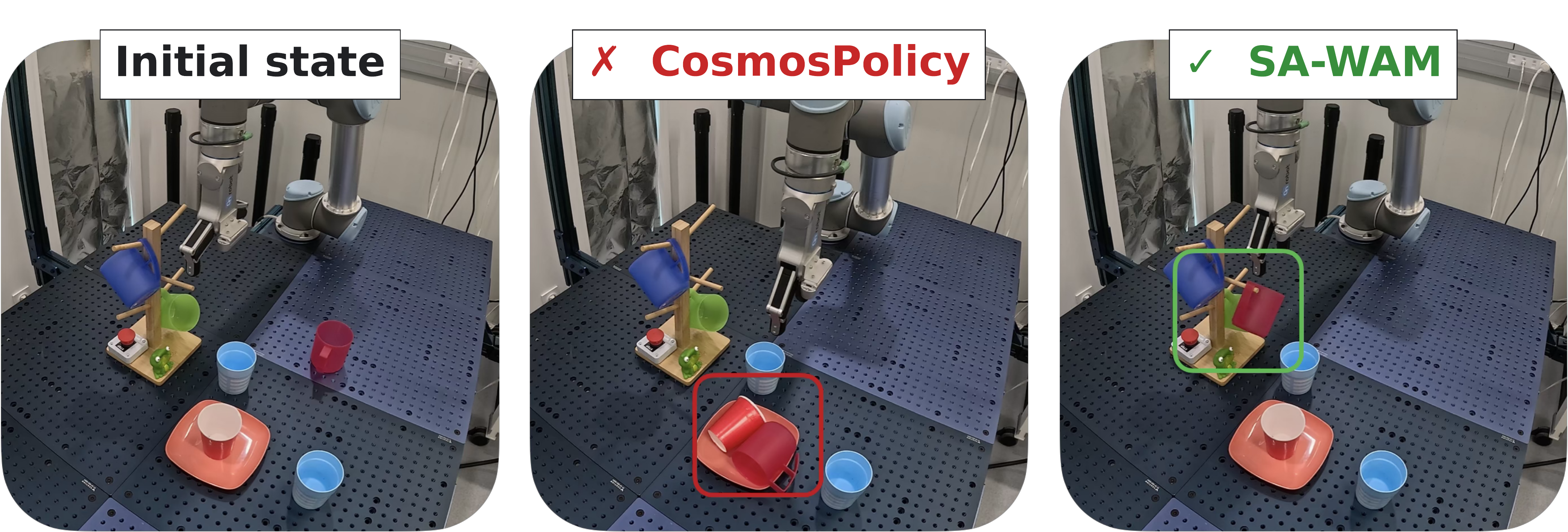}
 \caption{
\textbf{Task:}
\texttt{take the pink mug and put it on the middle part of the hanger}.
}
\label{fig:ur5_qual_green_mug}
\end{subfigure}
\vspace{1mm}
\begin{subfigure}{\columnwidth}
\centering
\includegraphics[width=\linewidth]{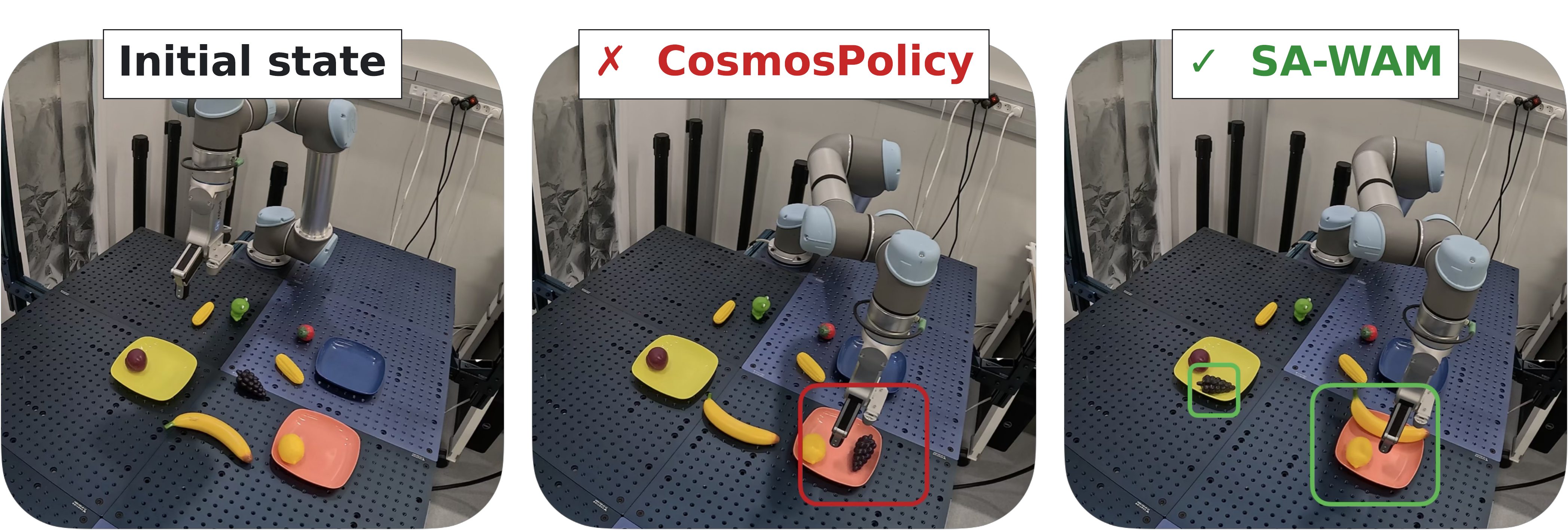}
\caption{
\textbf{Task}: \texttt{put the grapes in the yellow plate, then put the banana in the
pink plate}.
}
\label{fig:ur5_qual_grapes}
\end{subfigure}
\caption{\textbf{Qualitative examples in randomized environments for the UR5 experiments:} Two manipulation examples under domain randomization where SA-WAM reaches task completion while Cosmos-Policy fails to disambiguate visually similar distractors. We highlight rollout errors with red squares, and task completion with green squares.}
\label{fig:ur5_qualitative_examples}
\end{figure}

\noindent \textbf{Real-world UR5 setup.}
We evaluate SA-WAM in a real-world tabletop manipulation setting,
comparing it with Cosmos-Policy~\cite{kim2026cosmos} and
$\pi_0$~\cite{black2025pi0}. The platform consists of a 6-DoF UR5 arm
equipped with an RG6 parallel gripper, see Figure~\ref{fig:ur5_qualitative_examples}. The workspace is observed by a
fixed Orbbec Femto Mega RGB-D camera, which provides time-aligned RGB
images and metric depth. 

Robot states and actions are represented in end-effector (EE) space.
The state contains the current EE pose and gripper state, while each
action specifies an absolute EE waypoint and gripper command. Both are
represented as 8-D vectors,
$(x,y,z,q_x,q_y,q_z,q_w,\mathrm{gripper})$,
where $(x,y,z)$ denotes the EE position,
$(q_x,q_y,q_z,q_w)$ its orientation as a unit quaternion, and the final
scalar denotes the normalized gripper aperture.

Actions correspond to semantic key-step waypoints rather than dense
control-rate deltas or joint-space commands. The training dataset contains
20 demonstrations for each of 10 tasks, grouped into five categories: \texttt{single-step, put-in-plates, put-in-box, mug-to-hanger,} and \texttt{stack-cups}. For evaluation, we exclude the single-step
category, as it consists of isolated primitive behaviors that also appear
as subgoals within the composite tasks. Episodes contain
3--11 key steps, each providing an RGB-D observation together with the
corresponding 8-D EE state and absolute EE action.

\noindent \textbf{Training details.}
We train Cosmos-Policy and SA-WAM for 2,400 iterations on eight H100 GPUs with an effective batch size of 192. The learning-rate schedule consists of a 300-step warm-up followed by a linear decay to zero. The model predicts chunks of three key-step actions, of which the first two are executed at inference.

\noindent \textbf{UR5 results.}
Figure~\ref{fig:ur5_results} reports results on the four composite
manipulation categories mentioned above.
Each category is evaluated under clean and randomized conditions.
In the clean setting, object positions differ from the demonstrations,
while the remaining scene follows the training distribution. The
randomized setting additionally introduces distractor objects and visual
perturbations.

We perform 10 trials per category in each setting, corresponding to
20 trials per category overall. For multi-stage tasks, partial credit is
assigned according to task progress: each completed subgoal contributes
$0.5$ in two-stage tasks and $0.25$ in four-stage tasks. We compare
SA-WAM with Cosmos-Policy~\cite{kim2026cosmos} and
$\pi_0$~\cite{black2025pi0}.
For $\pi_0$, we initialize from the
official pretrained weights and fine-tune the full model on the same
key-step dataset with action horizon $H=3$, using the
LeRobot library~\cite{cadene2026lerobot}.

Figure~\ref{fig:ur5_results}(\subref{fig:ur5_clean}) shows the evaluation results in the clean setting. SA-WAM matches or exceeds both baselines in all
four categories, reaching an aggregate completion score of $90.0\%$,
compared with $75.0\%$ for Cosmos-Policy and $21.3\%$ for $\pi_0$.
Under randomization, SA-WAM retains $77.5\%$, whereas Cosmos-Policy and
$\pi_0$ decrease to $48.8\%$ and $11.3\%$, respectively, as shown in Figure~\ref{fig:ur5_results}(\subref{fig:ur5_rand}). This trend is
consistent with explicit depth improving robustness when appearance alone
is insufficient to distinguish the target from distractors.

Figure~\ref{fig:ur5_qualitative_examples} shows representative randomized-environment cases in which SA-WAM succeeds and Cosmos-Policy fails on the same episode. The scenes contain layout changes and visually confusing distractors. These examples are consistent with depth providing a complementary spatial cue when
appearance alone is ambiguous.

\section{Conclusion}
\label{sec:conclusion}

We introduced SA-WAM, a spatially aware World Action Model that
injects explicit 3D information into the DiT latent-frame sequence. By leveraging a
log-scale normalization, SA-WAM encodes depth into the input
range expected by a frozen VAE tokenizer, enabling geometry-aware
world-action modeling without a dedicated 3D encoder. SA-WAM achieves
state-of-the-art results on RoboCasa and LIBERO-Plus simulation benchmarks. Moreover, SA-WAM's strong performance transfers to the real-world setting with a UR5 robotic arm. SA-WAM improves RGB prediction quality over the Cosmos-Policy baseline on the RoboCasa benchmark,
highlighting the value of explicit spatial grounding for WAM-based robot policies.\\
{\bf Limitations.} Although our approach improves over state-of-the-art WAMs, the problem of predicting inconsistent futures and actions is still present.  Geometrically consistent training and verification are possible avenues for future improvement. Moreover, further work is needed to improve the inference efficiency of WAMs.

\section*{Acknowledgment}
This work was granted access to HPC resources of IDRIS under the
allocation AD011017145 made by GENCI.  It was funded in part by the French government under management of Agence Nationale de la Recherche as part of the “France 2030” program, reference ANR-23-IACL-0008 (PR[AI]RIE-PSAI project) and the ANR project VideoPredict (ANR-21-FAI1-0002-01). Cordelia Schmid would like to acknowledge the support by the K\"orber European Science Prize.  
The authors thank Peteris Kulits, Federica Spinola and Zeeshan Khan for their valuable contributions to this project.

\bibliographystyle{IEEEtran}
\bibliography{IEEEabrv,IEEEexample}

\end{document}